\documentclass[10pt,twocolumn,letterpaper]{article}

\usepackage{iccv}
\usepackage{times}
\usepackage{epsfig}
\usepackage{graphicx}
\usepackage{amsmath}
\usepackage{amssymb}
\usepackage{bbm}
\usepackage{algorithm}
\usepackage{algorithmic}
\usepackage{subcaption}
\usepackage{pifont}
\usepackage{authblk}
\newcommand{\myparagraph}[1]{\vspace{3pt}\noindent{\bf #1}}

\usepackage[pagebackref=true,breaklinks=true,letterpaper=true,colorlinks,bookmarks=false]{hyperref}

\iccvfinalcopy 

\ificcvfinal\fi

\begin{document}

\title{Natural Perturbed Training for General Robustness of Neural Network Classifiers}

\author{Sadaf Gulshad} 
\author{Arnold Smeulders}
\affil{UvA-Bosch Delta Lab,
University of Amsterdam, The Netherlands\\}
\maketitle
\ificcvfinal\thispagestyle{empty}\fi

\begin{abstract}
   We focus on the robustness of neural networks for classification. To permit a fair comparison between methods to achieve robustness, we first introduce a standard based on the mensuration of a classifier's degradation. Then, we propose natural perturbed training to robustify the network. Natural perturbations will be encountered in practice: the difference of two images of the same object may be approximated by an elastic deformation (when they have slightly different viewing angles), by occlusions (when they hide differently behind objects) or by saturation, Gaussian noise etc. Training some fraction of the epochs on random versions of such variations will help the classifier to learn better. We conduct extensive experiments on six datasets of varying sizes and granularity. Natural perturbed learning show better and much faster performance than adversarial training on clean, adversarial as well as natural perturbed images. It even improves general robustness on perturbations not seen during the training. For Cifar-10 and STL-10 natural perturbed training even improves the accuracy for clean data and reaches the state of the art performance. Ablation studies verify the effectiveness of natural perturbed training. 
\end{abstract}

\section{Introduction}
 Recent research in machine learning and computer vision shows that changes in the inputs of convolutional neural networks like blur or noise can drastically change the class predictions in the real world \cite{dodge2017study,azulay2018deep,recht2018cifar}.  Considering the importance of robustness against natural perturbations \cite{hendrycks2019benchmarking} proposed a benchmark consisting of a subset of Image net \cite {5206848} with corruptions applied to them. Although they introduced five severity levels for each type of perturbations, they do not standardize the effect of the perturbations for a fair, quantitative comparison among the different perturbations. Therefore, in this work instead of qualitative evaluation, we introduce a standardization procedure to permit a quantitative evaluation of robustness among alternative types of perturbations to train a network. 

Several methods for the robustness of neural networks against natural perturbations have been proposed in the literature \cite{yin2019fourier,rusak2020increasing,hendrycks2019using}. \cite{yin2019fourier} hypothesized that Gaussian noise and adversarial training helps against perturbations in the high frequency domain. \cite{rusak2020increasing} showed that by generating properly tuned Gaussian or speckle noise it is possible to generalize a network to unseen perturbations. In order to systematically enhance and study the robustness of neural networks against perturbations in this paper, we introduce a simple yet effective training procedure \textit{natural perturbed training}: the network is first trained for $n_1$ epochs on clean images followed by $n_2$ epochs on naturally perturbed versions of the same training images. Unlike previous methods, this method of training does not require architectural changes and it is not computationally expensive, while any type of natural perturbation could be used with it.

Concurrently, training methods have been introduced to achieve robustness against adversarial perturbations \cite{goodfellow2014explaining,miyato2018virtual,song2019robust}. To date, it is an open problem whether adversarial perturbations help making networks robust against natural perturbations and vice versa \cite{zhang2019interpreting, engstrom2019exploring}. \cite{zhang2019interpreting} showed that adversarial training helps to reduce the texture bias in neural networks, however \cite{engstrom2019exploring} showed that adversarial perturbations do not generalize to natural transformations like translations and rotations. Therefore, in this work after standardization permits a fair comparison between differently trained networks for robustness, we evaluate whether adversarial perturbations generalize to natural perturbations and the other way around.

There is also an open debate in literature \cite{tsipras2018robustness,zhang2019theoretically,su2018robustness} about the trade-off between robustness and accuracy of clean image classifiers when networks are robustified with adversarial training. We found that our natural perturbed training procedure does not lead to the large drop in the performance on clean images as adversarial training does. For Cifar-10 and STL-10 natural perturbation even helps to improve the accuracy to reach the state of the art performance  \cite{sosnovik2019scale}, without the high computational costs of adversarial training.

Standardization is also useful in the evaluation of robustified networks for unseen perturbations. In contrast to  \cite{hendrycks2019benchmarking,laugros2019adversarial,rusak2020increasing}, we learn the quantitative effect of the type of training for robustness also against \textit{unseen perturbations}.

Our contributions are:
(1) Where in literature a qualitative evaluation of robustness is often used, we propose standardization to permit quantitative evaluation for the comparison between two alternative training procedures aimed at achieving robustness.
  (2)  We introduce natural perturbed training which is computationally fast and shows better performance than adversarial training on clean, adversarial as well as natural perturbations.
 (3)  Natural perturbed training is demonstrated to improve the quantitative robustness of perturbations both seen and unseen during the training.
(4)  Natural perturbed training even improves the performance of classifiers in the absence of perturbations (without using more data and at almost no costs).

\section{Related Work}
\myparagraph{Natural Perturbations and Robustness.} 
In \cite{fawzi2015manitest,kanbak2018geometric} authors showed that neural networks are not even robust to translations and rotations. \cite{geirhos2017comparing} deduced that the performance of neural networks drops significantly as compared to humans with the increase of the signal-to-noise ratio of images. \cite{dodge2017study} also concluded that although neural networks are on par in performance with humans, they fail to perform well in the presence of perturbations like Gaussian noise or blur, which are easily handled by humans. Therefore, it is crucial to build robustness against such perturbations into the classification without degrading the performance on clean images, especially in such applications like autonomous driving and health. 

To promote the study of robustness against naturally occurring perturbations a few benchmarks have been proposed \cite{hendrycks2019benchmarking,hendrycks2020many,geirhos2017comparing}. Closely related to our work, in \cite{hendrycks2019benchmarking} the authors have introduced a large benchmark for natural perturbations, quite a few of which will be correlated \cite{laugros2021increasing}. In our work we selected six more or less independent types of natural perturbations covering the breadth of styles, see Figure \ref{fig:pert_sample}. In the reference, the authors have gone through the effort of defining five levels of severity for each type of perturbation. These levels are based on the visual effect but not standardized on their effect on the classification. As robustness is primarily aimed at the loss of classification performance, in this work at first we quantitatively standardize the comparison among differently trained networks to analyze the effect on their robustness. 
\begin{table}
\begin{center}
\begin{tabular}{|l | c | c | c | c }
\hline
Input & \shortstack{Our \\ standardization} & \shortstack{MSE\\ ($x_n$,$\zeta^t(x_n)$)}  \\
\hline \hline
Adversarial $\zeta^A(x_n)$  & 10.22 & 0.02 \\
Elastic $\zeta^E(x_n)$   & 10.60 & 54.31 \\
Occlusion $\zeta^O(x_n)$   & 10.24 & 199.73 \\
Gaussian Noise $\zeta^N(x_n)$   & 10.10 & 11.79 \\
Wave $\zeta^W(x_n)$   & 10.18 & 602.61  \\
Saturation $\zeta^S(x_n)$   & 10.4 & 269.71  \\
Blur $\zeta^B(x_n)$   & 10.51 & 18.2  \\
\hline
\end{tabular}
\end{center}
\caption{Significance of our standardization for Cifar-10. Our method is made consistent in the drop of classification performance across perturbations (with an arbitrary small deviation still remaining). The mean square error (MSE) between clean $x_n$ and perturbed $\zeta^t(x_n)$ images shows how large perturbations in the image may be before the same degradation in performance is achieved.}\label{table:stand}
\vspace{-0.5cm}
\end{table} 
 Table \ref{table:stand} shows the significance of our standardization method for fair comparison of robustness. When using the mean square error (MSE) between clean and perturbed images for standardization of perturbations, we see that the MSE shows a large variation in classification performance among different types of perturbations. Especially the MSE calculated for adversarial and other natural perturbations show different behavior. This is because adversarial perturbations are generated in order to misclassify an image while keeping the optical difference between clean and adversarial images to a minimum. We start from  standardizing according to their effect on classification accuracy. Hence, we drop the accuracy of the network by a constant value for each type of perturbation Table \ref{table:stand}. This enables the comparison among different perturbations and the robustness of classifiers. 
 
\begin{figure}
        \centering
\includegraphics[width=0.8\linewidth, trim=0 0 0 0, clip]{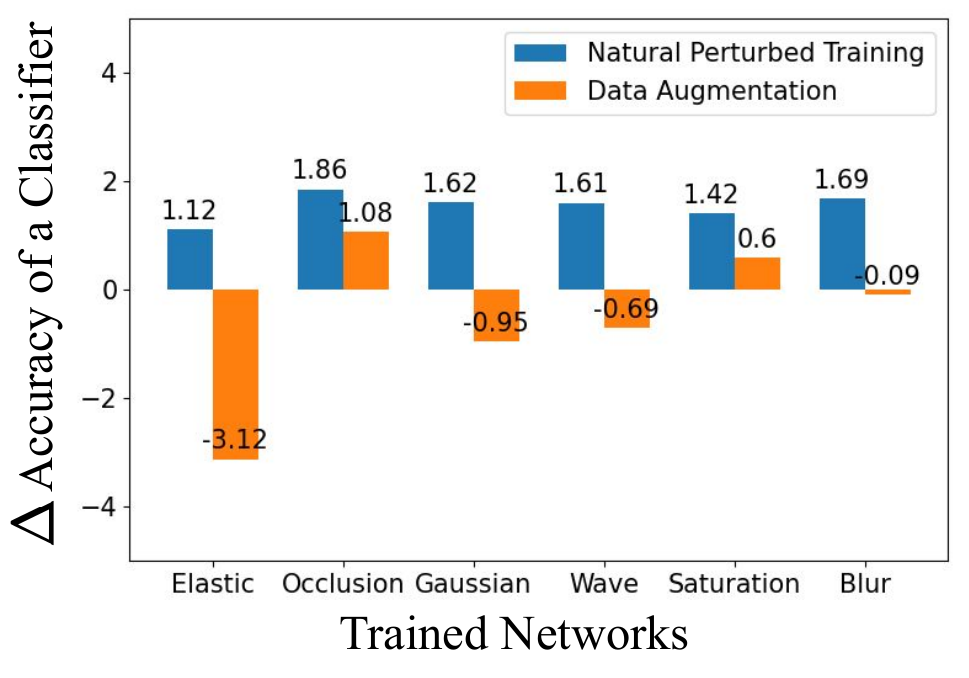}
    \caption{Comparing data augmentation with natural perturbed training on clean images for Cifar-10. $\bigtriangleup$ is the change in accuracy, positive values show improvement while negative values depict a drop in accuracy. We observe that natural perturbed training shows a better performance on clean images than data augmentation would. }
    \label{fig:aug-ours}
    \vspace{-0.5cm}  
      \end{figure}

Simultaneously, to improve the robustness against natural perturbations \cite{rusak2020increasing} performed data augmentation by carefully tuning Gaussian or speckle noise. \cite{tang2021selfnorm} introduced two normalization techniques SelfNorm and CrossNorm to enhance the generalization for out of distribution data. \cite{schneider2020improving} proposed to use batch normalization statistics calculated on corrupted images instead of clean images to enhance the robustness against perturbations. However, all the aforementioned approaches either require an extra network to find the suitable perturbation or a modification in the network. In contrast, in this paper, we introduce a training procedure in which after training on clean images we continue on perturbed versions of the clean inputs and minimize the loss for both of them. This leads to an improvement in the performance on robustness against perturbed images without requiring any architectural changes. Note that our natural perturbed training is different from standard data augmentation. Figure \ref{fig:aug-ours} contrasts the performance of clean images when the network is trained with the data augmentation versus when it is trained with natural perturbed training. For Cifar-10, we clearly see that natural perturbed training shows an improvement on clean image classification accuracy for all styles of perturbation. However, data augmentation either leads to a small  improvement or even a drop in the  performance with elastic, Gaussian and wave perturbations.

\myparagraph{Adversarial Perturbations and Robustness.} 
In \cite{szegedy2013intriguing}, the authors explored the robustness of neural networks. They showed that by adding small amounts of carefully crafted noise i.e. \textit{adversarial perturbations} to the images it is possible to change the prediction of the classifier. Since then plenty of research \cite{kurakin2016adversarial,papernot2016limitations,su2019one,carlini2017towards,moosavi2016deepfool} has been performed on finding different types of adversarial perturbations and study the robustification against them \cite{goodfellow2014explaining,kurakin2016adversarial,goodfellow2014explaining,carlini2017towards,dong2020benchmarking}. In this work we utilize a strong yet undefended attack basic iterative method \cite{kurakin2016adversarial} for generating adversarial perturbations and projected gradient descent \cite{madry2017towards} for adversarial training, which is one of the state of the art defense methods, for the comparison with natural perturbed training.

Although adversarial training helps to enhance the performance against adversarial perturbations, \cite{zhang2019theoretically,tsipras2018robustness} showed that with an increased robustness of adverserially trained neural networks in classification with such perturbations, simultaneously the network decreases in accuracy of the classification of clean images. This behavior deviates for our natural perturbed training. Apart from increasing the robustness for classification of perturbed images, the network retains its accuracy for clean images for most datasets, and even enhances its performance on CUB, StanfordCars, Cifar-10 and STL-10 datasets.

\cite{ford2019adversarial} established connections between adversarial and natural perturbation robustness, suggesting that neural networks should be robustified against both of them. \cite{rusak2020increasing} focused on robustification against adversarial as well as natural perturbations by using properly tuned Gaussian and Speckle noise. In this work, instead of generating tuned noise and then training the network, we refrain from tuning noise during training. We show that our natural perturbed training shows better performance with occlusion, elastic and wave than with Gaussian noise as perturbation.

\section{Methods}
Given the $n^{th}$ input image $x_n$ and its respective output $y_n$, a classifier $f$ predicts the class $f(x_n)=y_n$. Here we consider the problem of robust classification against artificially created adversarial $\zeta^A$ and natural $\zeta^t$ perturbations as noise, motion blur, difference in viewing angle, color saturation, and occlusion.

\subsection{Quantitative Standardization}
As the evaluation metric for classification is accuracy, we propose to add perturbations in the input images such that the performance drop $\rho$ in classification accuracy is equal for all perturbations under consideration as shown in the Table \ref{table:stand}. It is given as:
\begin{align}
    \resizebox{\linewidth}{!}{$\rho = \left[\frac{1}{n} \sum_{n=1}^N\mathbbm{1}(f(x_n)=y_n)\right]-\left[\frac{1}{n} \sum_{n=1}^N\mathbbm{1}(f(\zeta^t(x_n))=y_n)\right]$}
    \label{eq:1}
\end{align}
where $\mathbbm{1}$ is the indicator function. Hence, we set the parameters of each $\zeta^t$ under consideration such that the drop $\rho$ is constant for each type of perturbation.
\subsection{Perturbations} \label{nat_pert}
\begin{figure}[t]
        \centering
\includegraphics[width=0.9\linewidth, trim=0 0 0 0, clip]{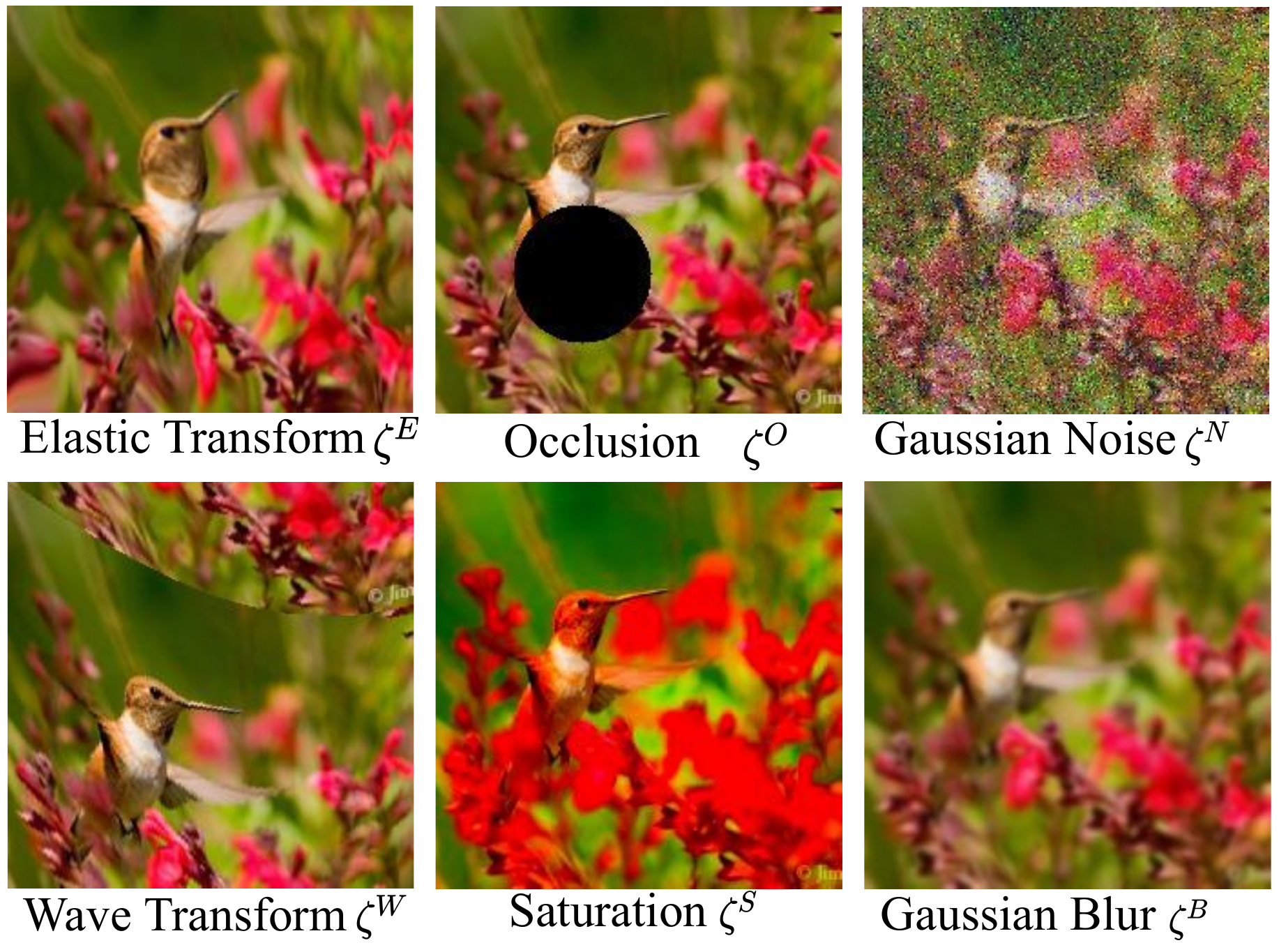}
    \caption{Randomly selected sample images for the six natural perturbations under consideration in our experiments.}
    \label{fig:pert_sample}
    \vspace{-0.5cm}  
\end{figure}

\myparagraph{Natural Perturbations.} 
We consider a set of natural perturbations $\zeta^t$ with least correlations among them, where $t\in \{E,O,N,W,S,B\}$ represents the type of perturbation operator. We create perturbed images by selecting a perturbation from $t$ and applying it on the image $\zeta^t(x_n)$. This leads to a drop in the performance of the classifier $f(\zeta^t(x_n))$. Samples for the six natural perturbations under consideration are shown in Figure \ref{fig:pert_sample}.

The first natural perturbation is elastic deformation $\zeta^E$. Elastic deformation usually appears in small variations in the viewing angle of the recording. Similar to \cite{simard2003best} we add elastic deformation to the images by applying $\zeta^E= \mathcal{T}(x_n,\alpha x'_n\circledast{\mathcal{N}(\mu, \sigma^2)})$ on the image. We generate random displacement fields by selecting a random number between $-1$ and $+1$ i.e. $x'\in rand(-1,+1)$. Then we apply a Gaussian filter by convolving these fields with it i.e. $\alpha x'_n\circledast{\mathcal{N}(\mu, \sigma^2)}$, where $\alpha$ is the intensity of deformation and $\mathcal{T}$ is the affine transform. Occlusions are created by selecting minimum values from $\zeta^O= \text{min} (x_n,b(x_c ,t,r))$, where, $b$ is a matrix of zeros with $x_c$ as its center and $t$, $r$ being the thickness and radius of the circle respectively. Gaussian noise is introduced by $\zeta^N (x_n)= x_n + {N(\mu, \sigma^2)}$. A wave transform is added by $\zeta^W= x_n \longmapsto (Sin(2\pi x_n w))$, where $\longmapsto$ is the roll operator (in numpy) which rolls the original image by $Sin(2\pi x_n w)$. Saturation is introduced by using $\zeta^S= (1-\alpha)x'+\alpha x_n$, where, $\alpha \in [0,1]$, $x'$ is the black and white version of $x_n$. Gaussian blur $\zeta^B$ is introduced by convolving a two-dimensional Gaussian function to the image.

The natural perturbations are class agnostic in a stochastic sense. However, they are made image specific by selecting different perturbations for different images. For elastic deformation, we vary the intensity of elasticity for each image such that it leads to a specific drop $\rho$. For occlusion, the position of occlusion is randomly selected for each image, the intensity of Gaussian noise is also randomly uniformly varied. Per image, the wave is scaled uniformly at random, as are the saturation factor and variance of the Gaussian blur filter.

\myparagraph{Adversarial Perturbations.}
Adversarial examples are generated while satisfying two properties 1) the class of the adversarial image is different from the class predicted for clean image i.e. $f(\zeta^A(x_n))\neq f(x_n)$, 2). Perturbed and original images are visually similar and their similarity is determined by the $l_p$-norm. While fulfilling these two properties we use a basic iterative method \cite{kurakin2016adversarial} for generating adversarial examples $\zeta^A(x_n)$. We find the perturbation $\delta_n$ with a small norm $l_\infty$ bounded by $\epsilon$ such that $f(x_n)\neq f(\zeta^A(x_n))$, where $\zeta^A(x_n)=x_n+\delta_n$ and $\delta_n \leq \epsilon$. We solve the following equation:
\begin{align}
    \zeta^A(x_n^0)&=x_n+\delta \\
    \zeta^A(x_n^{k+1})&=\text{Clip}_\epsilon \{\zeta^A(x_n^k)+ \\  &\epsilon_s Sign(\bigtriangledown_{x}(\mathcal{L}_r^{\delta}(\zeta^A(x_n^k),y_n,\theta))\}
\end{align}
where $\mathcal{L}_r^{\delta}(\zeta^A(x_n^k),y_n,\theta)$ represents the gradient of cost function w.r.t the perturbed image $\zeta^A(x_n^k)$ at step $k$, $\epsilon_s$ determines the step size taken in the direction of sign of the gradient and the result is clipped by $\epsilon$.

\subsection{Robustness}
The neural network classifier is trained by minimizing the loss function: 
 \begin{align}
    \mathcal{L}_s=\underset{\theta}{\text{min}}\, \frac{1}{|S|}\sum_{(x_n,y_n)\in S}\mathcal{{L}}(f(x_n),y_n)
    \label{eq:4}
\end{align}
where $S = \{(x_n, y_n)| x_n \in X, y_n \in Y \}$ is the training set, $\theta$ the network parameters and $\mathcal{L}$ the cross-entropy loss. Usually, the data augmentation is performed by adding perturbed versions of the input images. The network is trained by replacing the clean input image $x_n$ with its perturbed version $\zeta^t(x_n)$ in Equation \ref{eq:4}. 
\begin{figure}[t]
        \centering
\includegraphics[width=\linewidth, trim=0 0 0 0, clip]{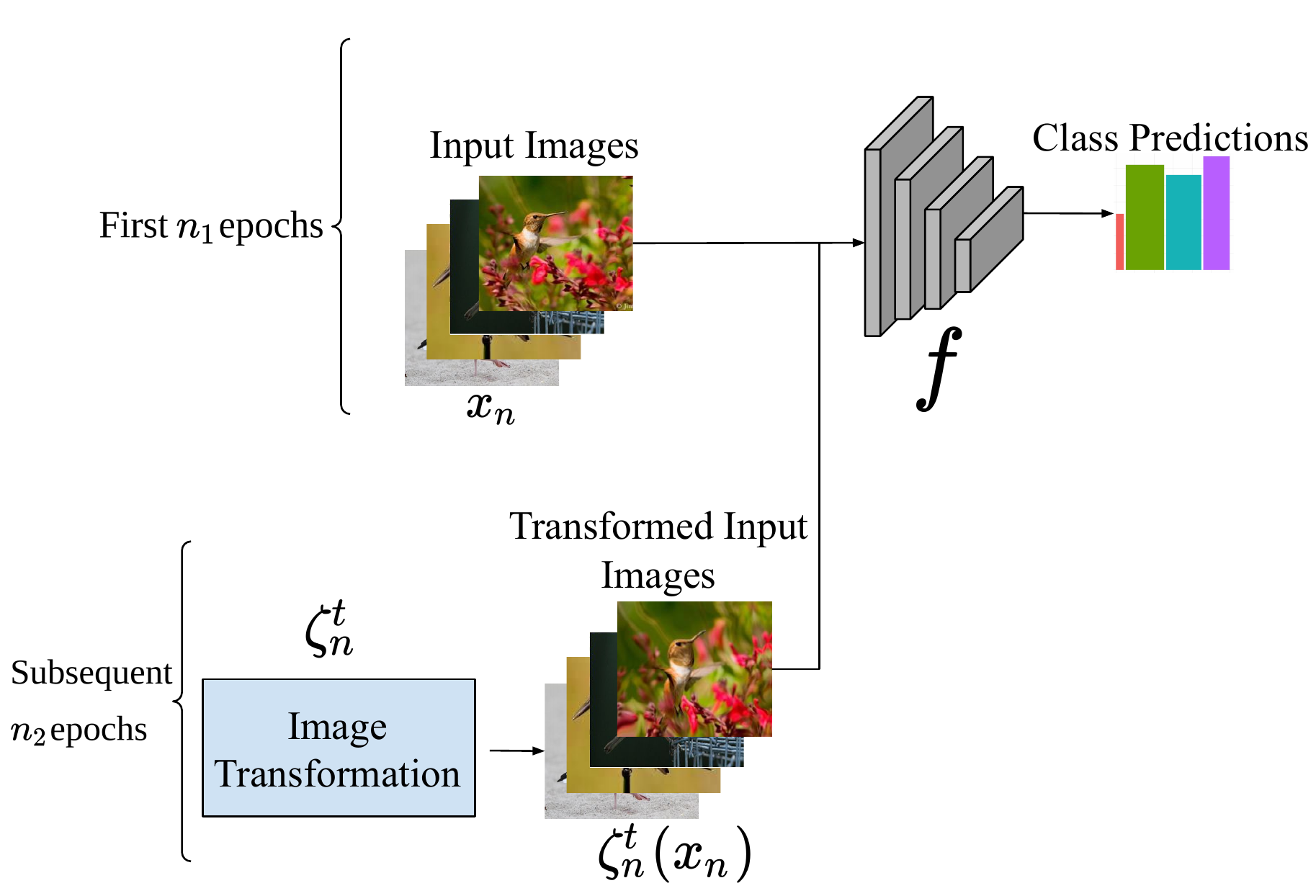}
    \caption{Our natural perturbed training procedure. We train a network $f$ on clean samples $x_n$ for first $n_1$ epochs. In subsequent $n_2$ epochs we add perturbed versions of input images $\zeta^t(x_n)$ in the training while optimizing the loss for both clean and perturbed samples for the rest of epochs. }
    \label{fig:train_proc}
    \vspace{-0.5cm}  
\end{figure}
      
\myparagraph{Natural Perturbation Robustness.}
In order to learn better loss surfaces for clean image classification and robustification against perturbed inputs, in this work we introduce \textit{natural perturbed training} as shown in Figure \ref{fig:train_proc}. We start training the classifier with clean images $x_n$ for $n_1$ epochs while optimizing the loss $\mathcal{L}_s$. Then we add their perturbed versions $\zeta^t(x_n)$ besides the clean for the subsequent $n_2$ epochs while minimizing the loss for both of them i.e. $\mathcal{L}_r^{\zeta}=\frac{\mathcal{L}_s+\mathcal{L}^{\zeta}}{2}$, where $\mathcal{L}^{\zeta}$ is the loss for perturbed samples. The procedure for natural perturbed training is given in the box \ref{alg:nat-pert-train}.

\myparagraph{Adversarial Robustness.} 
For adversarial robustness, we consider adversarial training as described in \cite{goodfellow2014explaining}. The network is trained on adversarial samples besides clean images, while the loss function is optimized for both clean and adversarial samples given by: 
\begin{align}
\mathcal{L}^{\delta}&=\underset{\theta}{\text{min}}\, \frac{1}{|S|}\sum_{(\zeta^A(x_n),y_n)\in S}\mathcal{{L}}(f(\zeta^A(x_n)),y_n)\\
    \mathcal{L}_r^{\delta}&={\mathcal{L}_s+\mathcal{L}^{\delta}}
\end{align}
where $\mathcal{L}_s$ is the loss for clean images and $\mathcal{L}^{\delta}$ is the loss for adversarial images.

\begin{algorithm}[H]
\begin{algorithmic}[1]
\STATE  Given $S = \{(x_n, y_n)| x_n \in X, y_n \in Y \}$, learning rate $\eta$ and a set of natural perturbations $\zeta^t$. 
\STATE Initialize $\theta$ randomly

\FOR{$epoch=1$ to $n_1+n_2$}
\FOR{$minibatch$  $B\subset|S|$}
      			
    \STATE $\mathcal{L}_s=\mathcal{{L}}(f(x_n),y_n,\theta)$
    \IF{$epoch > n_1 $} 
    \STATE $\mathcal{L}^{\zeta}=\mathcal{L}(f(\zeta^t(x_n)),y_n,\theta)$   
    \STATE $\mathcal{L}_r^{\zeta}=\frac{\mathcal{L}_s+\mathcal{L}^{\zeta}}{2}$
        		\ENDIF
        \STATE Update $\theta$ with SGD.
        \STATE $\theta=\theta-\eta\bigtriangledown_{\theta}\mathcal{L}_r^{\zeta}$
        		
\ENDFOR
\ENDFOR
\end{algorithmic}
\caption{Natural Perturbed Training for Robustification.}
\label{alg:nat-pert-train}
\end{algorithm}
\subsection{Implementation Details}
\myparagraph{Evaluation Metric.} We use change in the accuracy $\bigtriangleup$ as the evaluation metric for the robustness of classifiers. The change is calculated between a standard classifier for clean inputs $f(x_n)$ and a robustified classifier for clean $f_r(x_n)$ or perturbed $f_r(\zeta^t(x_n))$ inputs. The change in the accuracy is given by: 
\begin{align}
    \resizebox{\linewidth}{!}{$\bigtriangleup = \left[\frac{1}{n} \sum_{n=1}^N\mathbbm{1}(f(x_n)=y_n)\right]-\left[\frac{1}{n} \sum_{n=1}^N\mathbbm{1}(f_r(\zeta^t(x_n))=y_n)\right]$}
    \label{eq:8}
\end{align}
where $\mathbbm{1}$ is the indicator function.  

\myparagraph{Standard Network Training and Testing.} We perform classification using Resnet-152. For Cifar-10 we train the networks from scratch. For other datasets networks are pre-trained on Image-net and fine-tuned on the respective datasets. The networks are tested for both clean and perturbed inputs. Natural perturbations are generated using the method described in section \ref{nat_pert} while keeping the drop $\rho$ from equation \ref{eq:1} the same for all perturbations to ensure standardization. To make the perturbations diverse across each image we select the parameters of perturbations randomly. Adversarial perturbations are created using the basic iterative method with the number of steps $K$ taken as $10$ and $\epsilon$ values such that the drop $\rho$ is the same as for other perturbations. The metric of similarity between clean and adversarial samples is $l_{\infty}$ norm.

\myparagraph{Robust Network Training and Testing.} Networks are robustified with natural perturbed training, see the box \ref{alg:nat-pert-train}. Each network is robustified with one type of perturbation and the parameters for perturbations are tuned such that they lead to a constant drop $\rho$, see Equation \ref{eq:1}. Adversarial training is performed using projected gradient descent (PGD) with $K=10$ and $\epsilon$ tuned such that it leads to the same drop $\rho$ as the drop of other perturbations. The parameters for the optimizer, learning rate scheduler, and number of epochs are constant across adversarial training and natural perturbed training within a dataset. PGD adversarial training makes $O(KS)$ computational gradient steps in one epoch where $K$ is the number of steps and $S$ is the dataset size. This procedure is $K$ times slower than the standard training $O(S)$ \cite{wong2020fast} hence, our perturbed natural training is equally faster than adversarial training. 

\section{Experiments and Results}
We compare natural perturbed training with adversarial training on clean, natural perturbed and adversarial inputs. In all plots, a symbol represents one run on a trained network with one specifically (perturbed or clean) test set: the symbol represents the test perturbation type while the color represents the training perturbation type.
\begin{figure}[t]
        \centering
\includegraphics[width=0.8\linewidth, trim=0 0 0 0, clip]{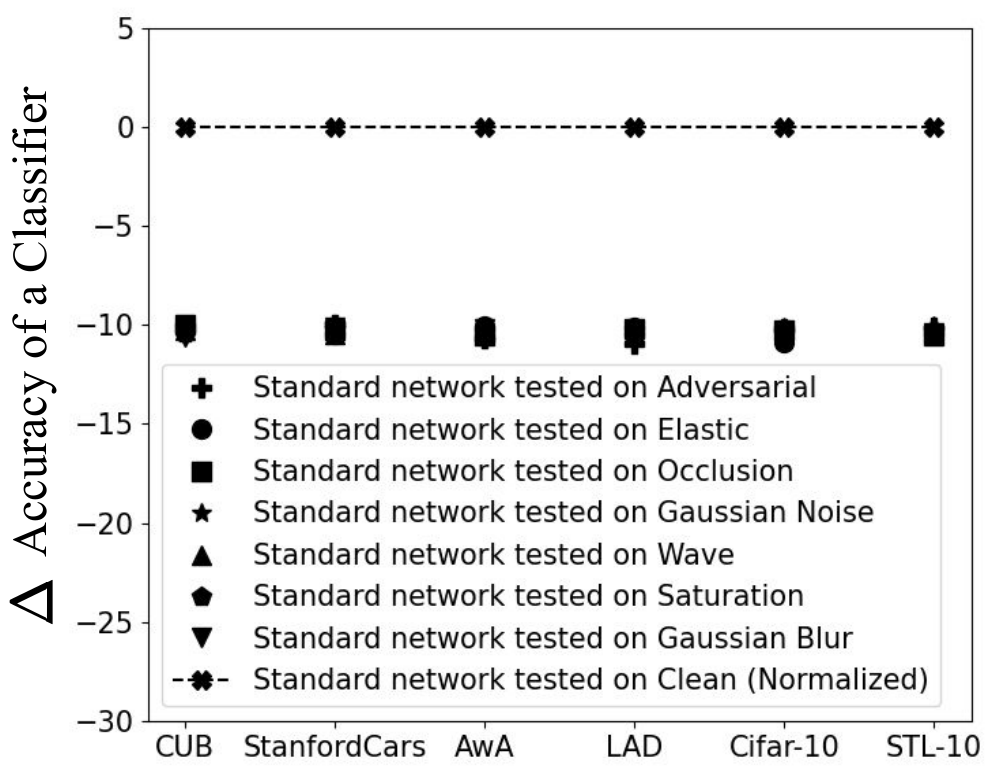}
    \caption{Standardization by calibrating the drop $\rho$. The cross symbol at zero shows the normalized accuracy of a standard network for clean images. Each of the symbols on $-10$ shows the standardization by dropping the performance of a standard network when a perturbation is introduced. Hence, the overlap of symbols at $-10$ for all the perturbations show the degree to which the standardization is uniform.}
    \label{fig:drop}
    \vspace{-0.5cm}  
\end{figure}
\begin{figure*}[t]
     \centering
     \begin{subfigure}[b]{0.4\textwidth}
         \centering
         \includegraphics[width=\textwidth]{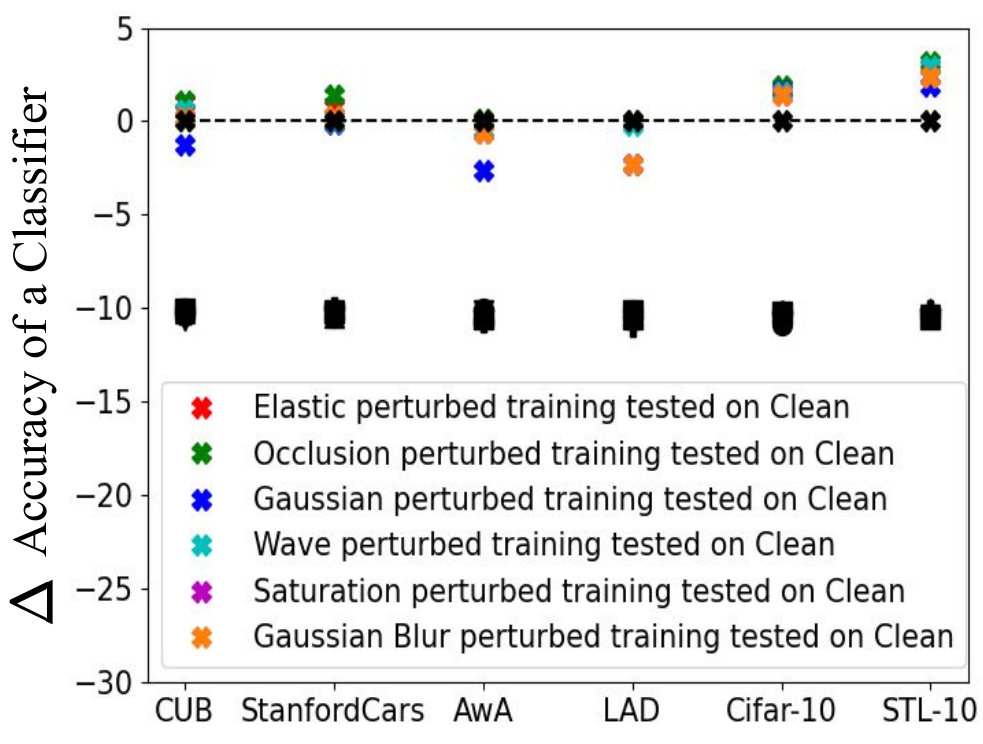}
         \caption{{Evaluating Natural perturbed training for clean images.} }
         \label{fig:nat_clean}
     \end{subfigure}
     \hspace{1em}%
     \begin{subfigure}[b]{0.4\textwidth}
         \centering
         \includegraphics[width=\textwidth]{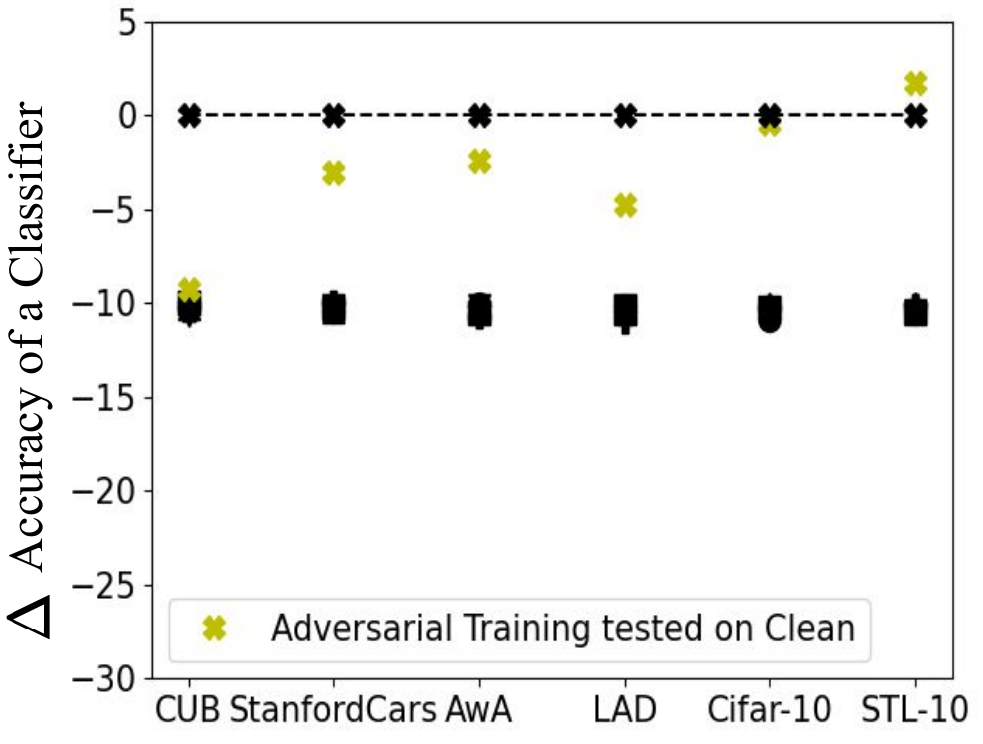}
         \caption{{Evaluating Adversarial training for clean images\\.  }  }
         \label{fig:AT_clean}
     \end{subfigure}
        \caption{Comparing the performance of natural perturbed training with adversarial training for clean images, where the cross symbol represents a clean test set and the color of the symbol represents the type of training perturbation. Adversarial training degrades the accuracy in the classification of clean images, but natural perturbed training does not degrade the performance on clean images. It even improves the classifier accuracy for four in six datasets.}
        \label{fig:clean}
        \vspace{-0.5cm}
\end{figure*}

\myparagraph{Datasets.}
Six datasets of varying granularity and size are used in our experiments. Cifar-10 \cite{krizhevsky2009learning} consists of ten coarse-grained classes with 50000 training and 10000 test images. STL-10 \cite{coates2011analysis} contains 5000 training and 8000 test images belonging to ten coarse-grained categories. Different from Cifar-10 the image size is $96\times96$ pixels. The Large attribute dataset (LAD) \cite{zhao2018large} contains 78017 images with 230 fine-grained classes. We use 11702 training, 9947 validation and 9284 test images for our experiments. Animals with attributes (AwA) \cite{8413121} consists of 37322 images with 50 fine-grained classes. We use 10450 of them for training, 7524 for validation, and 9674 for testing. StanfordCars \cite{KrauseStarkDengFei-Fei_3DRR2013} contains 8144 train and 8041 test images with 196 fine-grained categories of cars. The CUB-birds dataset \cite{WelinderEtal2010} consists of 11788 images with 5395 for training, 599 for validation and 5794 for testing, divided over 200 fine-grained categories of birds. The input size for all fine-grained datasets is taken as $224\times224$ pixels.
\subsection{Standardizing Network Robustness}
\myparagraph{Normalizing Accuracy.} We begin by evaluating the performance of a standard neural network classifier for clean images. A standard classifier shows the test accuracy of $93.18$ for Cifar-10, $88.60$ for STL-10, $87.86$ for LAD, $84.79$ for AwA, $86.48$ for StanfordCars, and $81.20$ for CUB dataset. The performance of the standard classifiers for clean images is the reference value of zero, as indicated by the cross symbol, see Figure \ref{fig:drop}.

\myparagraph{Standardization by Calibrating the Drop $\rho$.} While considering the standard networks as the baseline, we standardize the comparison among robustness of different networks by setting the desired drop $\rho$ in Equation \ref{eq:1} at $10\%$ for each dataset, shown in Figure \ref{fig:drop} at $-10\%$. We succeed in reaching a standardized drop with a maximum deviation of $0.26\%$. Hence, our standardization enables fair comparison among robustified networks on different types of perturbations.
\subsection{Evaluating Robustified Networks on Clean Images}
We contrast the performance of adversarial training with natural perturbed training on the clean test set. Figure \ref{fig:nat_clean} shows the performance of a network trained with natural perturbed training and tested on clean inputs. Except for Gaussian blur on LAD and Gaussian noise on AwA and CUB, natural perturbed training retains the performance of the classifier on clean images. For CUB, StanfordCars, Cifar-10 and STL-10 datasets training with the perturbed natural images even leads to an improvement in performance as compared to a standard network trained only on clean images. We achieve a maximum of $95.04$ for Cifar-10 and $91.81$ for STL-10 with our natural perturbed training. 
\begin{figure*}[t]
     \centering
     \begin{subfigure}[b]{0.4\textwidth}
         \centering
         \includegraphics[width=\textwidth]{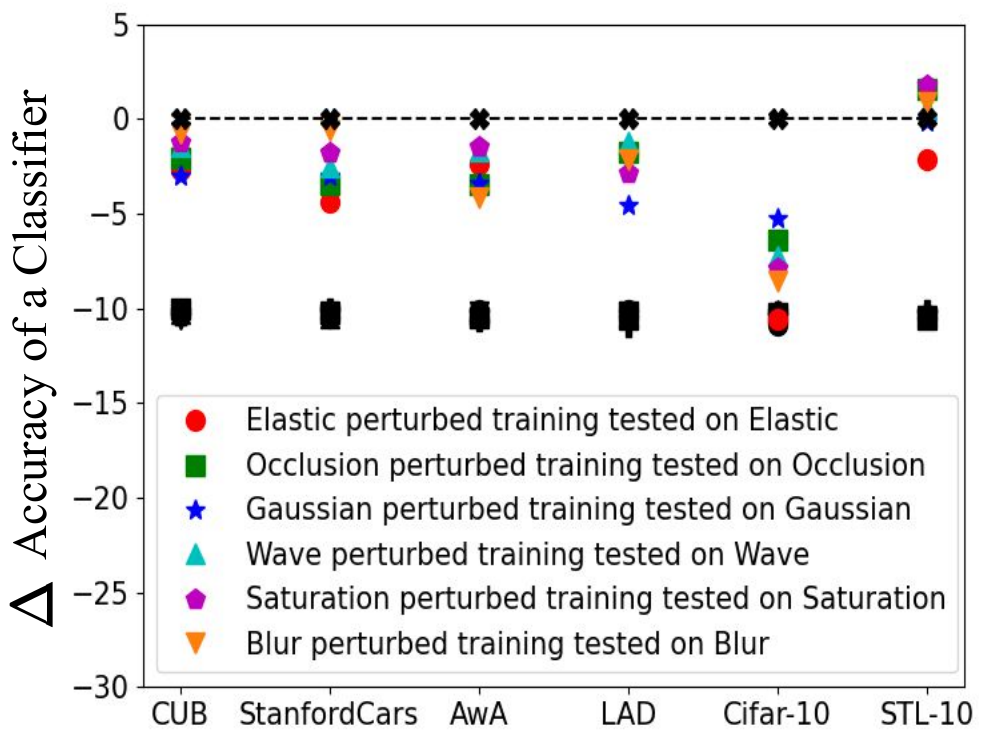}
         \caption{{Evaluating Natural perturbed training for seen natural perturbations. } }
         \label{fig:nat_seen}
     \end{subfigure}
     \hspace{1em}%
     \begin{subfigure}[b]{0.4\textwidth}
         \centering
         \includegraphics[width=\textwidth]{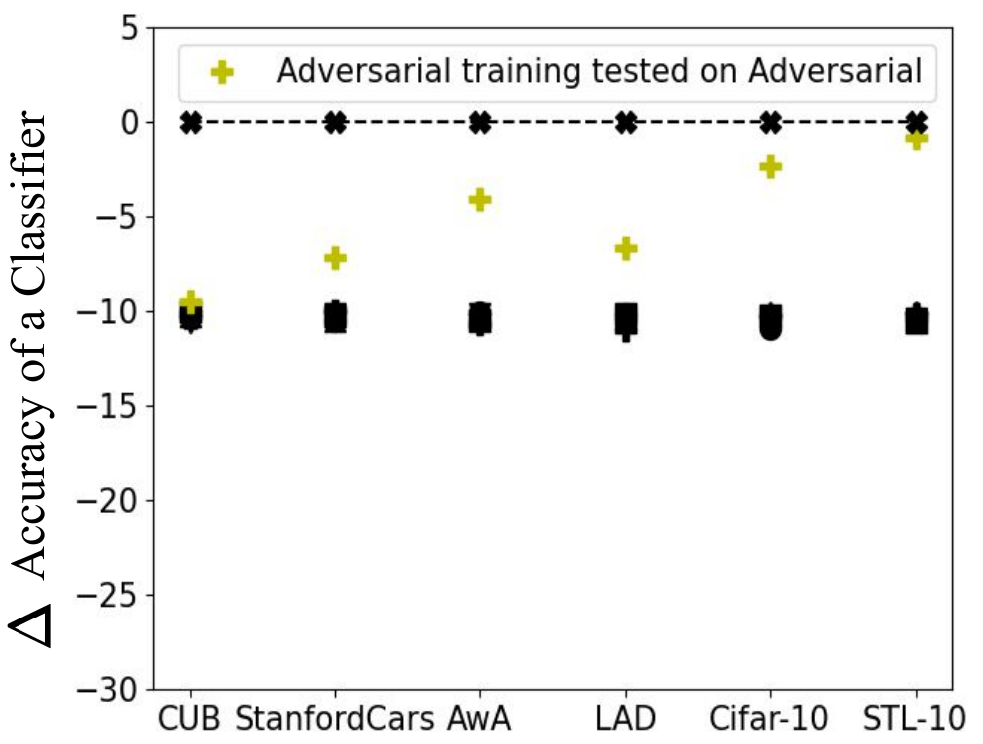}
         \caption{{Evaluating Adversarial training (AT) for adversarial perturbations.} }
         \label{fig:AT_adv}
     \end{subfigure}
        \caption{Comparing the performance of natural perturbed training with adversarial training on \textit{seen perturbations}. Where the type of the symbol represents the test perturbation type and color of the symbol represents the type of training perturbation. Adversarial training recovers the performance on adversarial images, but the recovery for natural perturbations with natural perturbed training is higher. }
        \label{fig:seen}
        \vspace{-0.5cm}
\end{figure*}
Figure \ref{fig:AT_clean} shows the performance of adversarially robustified networks on clean images. We see that robustifying networks against adversarial perturbations leads to the drop in the performance on clean images for all datasets except STL-10. Hence, adversarial training shows a trade-off between robustness on adversarial perturbations and clean image accuracy. In contrast, our natural perturbed training does not degrade clean image accuracy but leads to an improvement in the performance.

\subsection{Evaluating Robustified Networks on Seen Perturbations}
We evaluate the robustness of natural perturbed training on the same type of perturbation e.g. a network trained with elastic perturbed training tested on elastic (seen perturbations) as shown in Figure \ref{fig:nat_seen}. Results show that natural perturbed training helps to recover the performance when tested on seen perturbations for both coarse and fine-grained datasets. The recovery is highest for STL-10 and least for Cifar-10. Where Cifar-10 and STL-10 are both coarse-grained, the input size in Cifar-10 is around three times smaller than STL-10. Hence we argue that after introducing natural perturbations, the damage in Cifar-10 is too much to recover from. In general, all datasets show significant recovery in the performance with the natural perturbed training in the presence of seen perturbations. 

\begin{figure*}[t]
     \centering
     \begin{subfigure}[b]{0.31\textwidth}
         \centering
         \includegraphics[width=\textwidth]{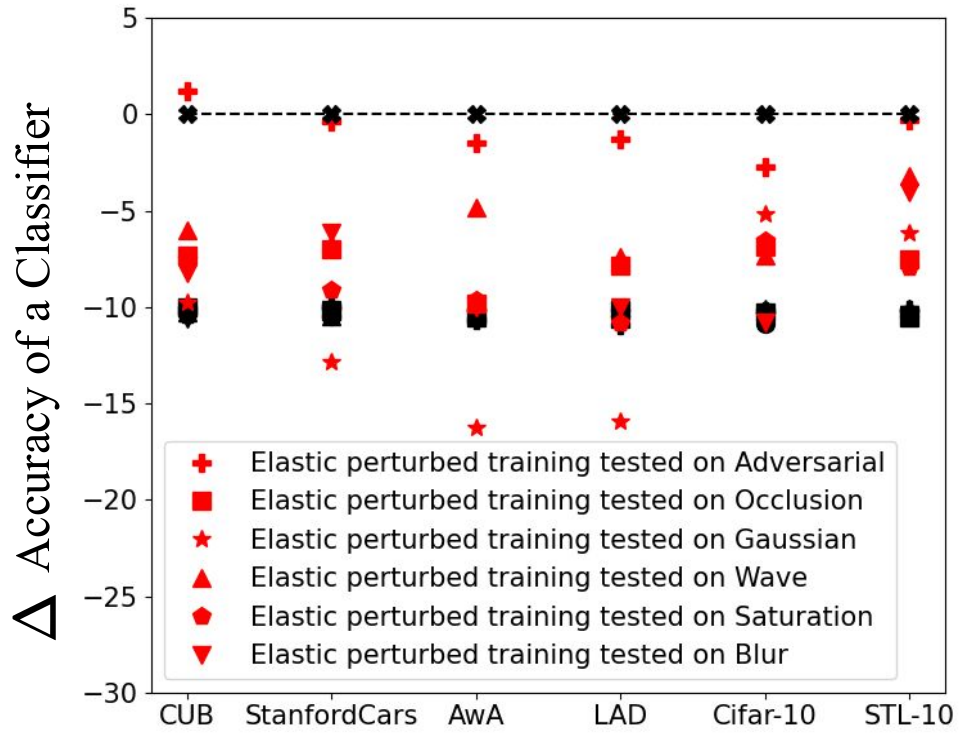}
         \caption{Evaluating elastic perturbed training for unseen natural perturbations.  }
         \label{fig:nat_elastic}
     \end{subfigure}
     \hspace{1em}%
     \begin{subfigure}[b]{0.31\textwidth}
         \centering
         \includegraphics[width=\textwidth]{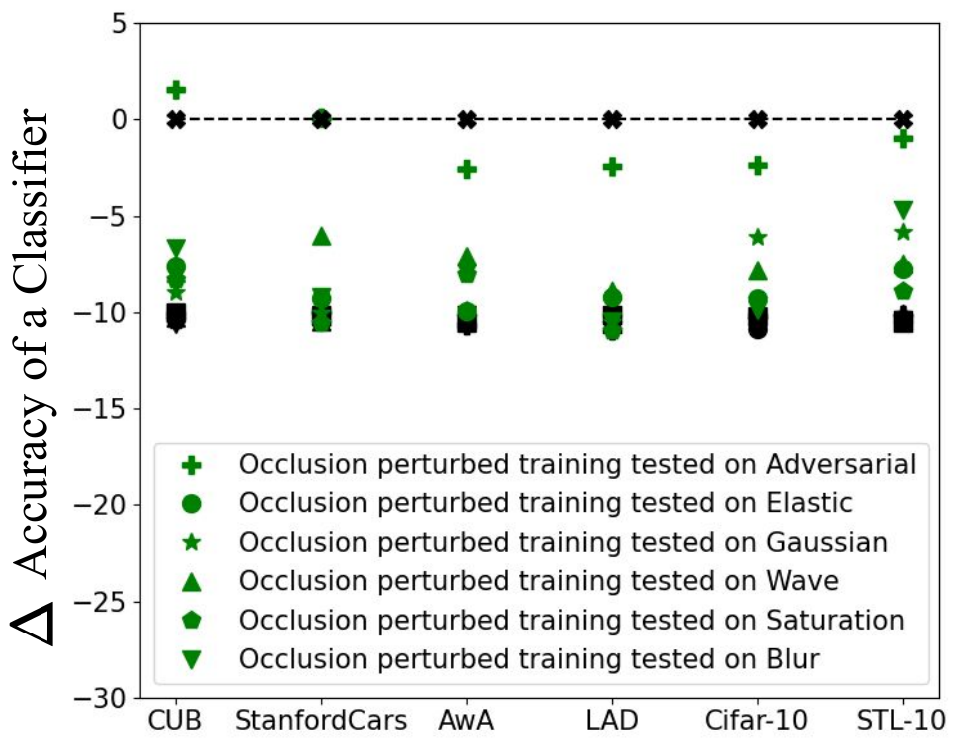}
         \caption{Evaluating occlusion perturbed training for unseen natural perturbations.  }
         \label{fig:nat_occ}
     \end{subfigure}
     \hspace{1em}%
     \begin{subfigure}[b]{0.31\textwidth}
         \centering
         \includegraphics[width=\textwidth]{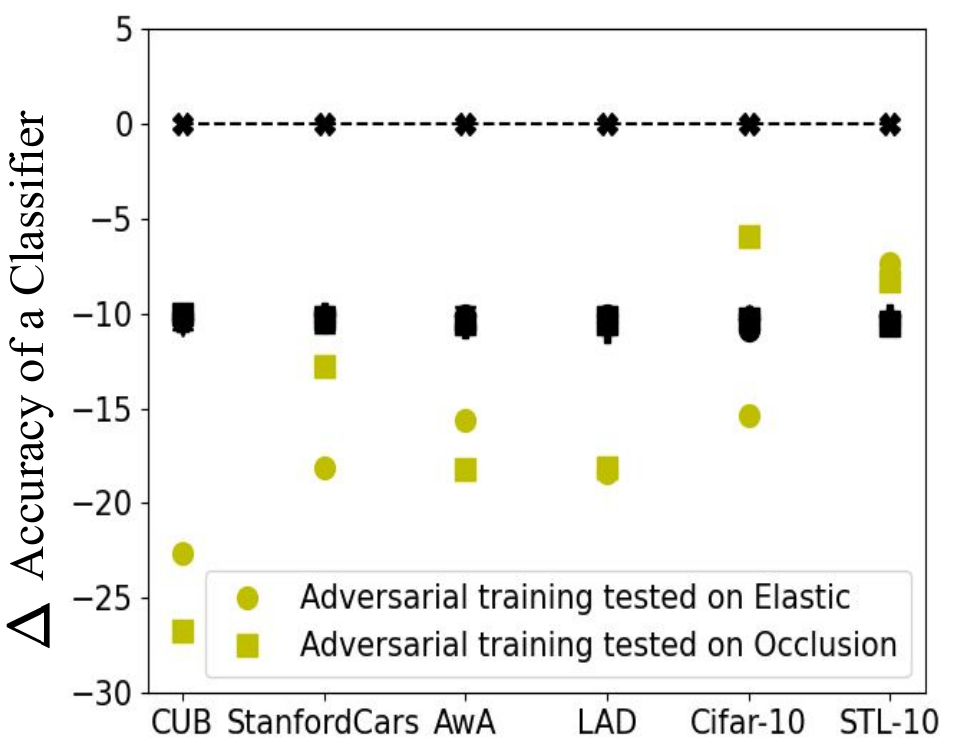}
         \caption{{Evaluating Adversarial training for unseen natural perturbations.  } }
         \label{fig:AT_nat}
     \end{subfigure}
        \caption{Comparing the performance of Natural perturbed training with Adversarial training on \textit{unseen perturbations}. The type of symbol represents test perturbation and color of the symbol represents the type of training perturbation. Adversarial training shows some general robustness on coarse-grained datasets but for fine-grained datasets it fails to generalize. Natural perturbed training generalizes to adversarial perturbations and other natural perturbations.}
        \label{fig:unseen}
        \vspace{-0.4cm}
\end{figure*}
\begin{figure*}[t]
     \centering
     \begin{subfigure}[b]{0.31\textwidth}
         \centering
         \includegraphics[width=\textwidth]{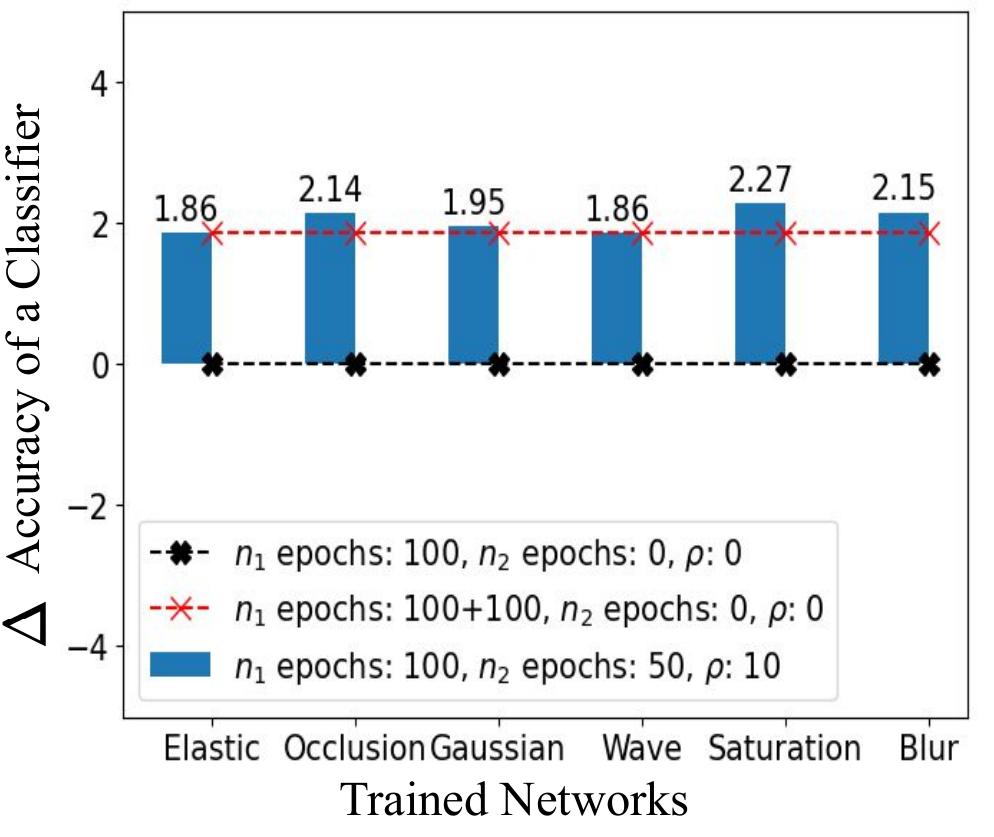}
         \caption{Standard training for a large number of epochs vs natural perturbed training for less epochs i.e  $\#(n_1+n_1)>$ $\#(n_1+n_2)$.}
         \label{fig:vary_clean}
     \end{subfigure}
     \hspace{1em}%
     \begin{subfigure}[b]{0.31\textwidth}
         \centering
         \includegraphics[width=\textwidth]{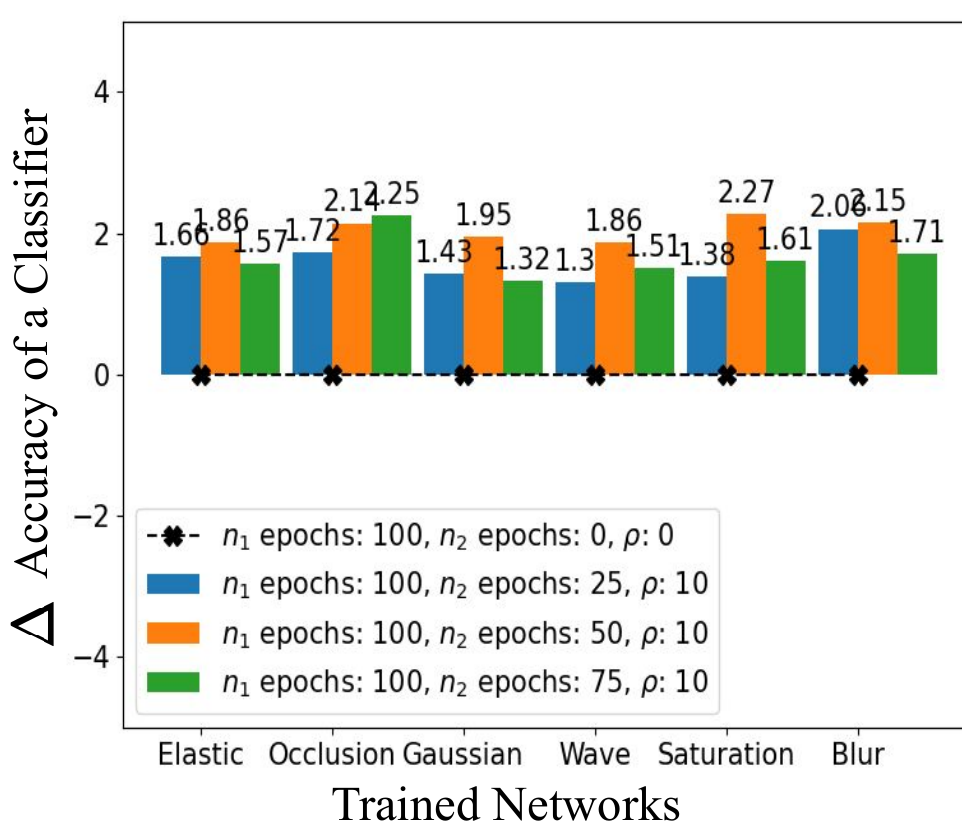}
         \caption{Evaluation by varying the number of subsequent ($n_2$) epochs to 25, 50 and 75 \\.}
         \label{fig:vary_pert_epoch}
     \end{subfigure}
     \hspace{1em}%
     \begin{subfigure}[b]{0.31\textwidth}
         \centering
         \includegraphics[width=\textwidth]{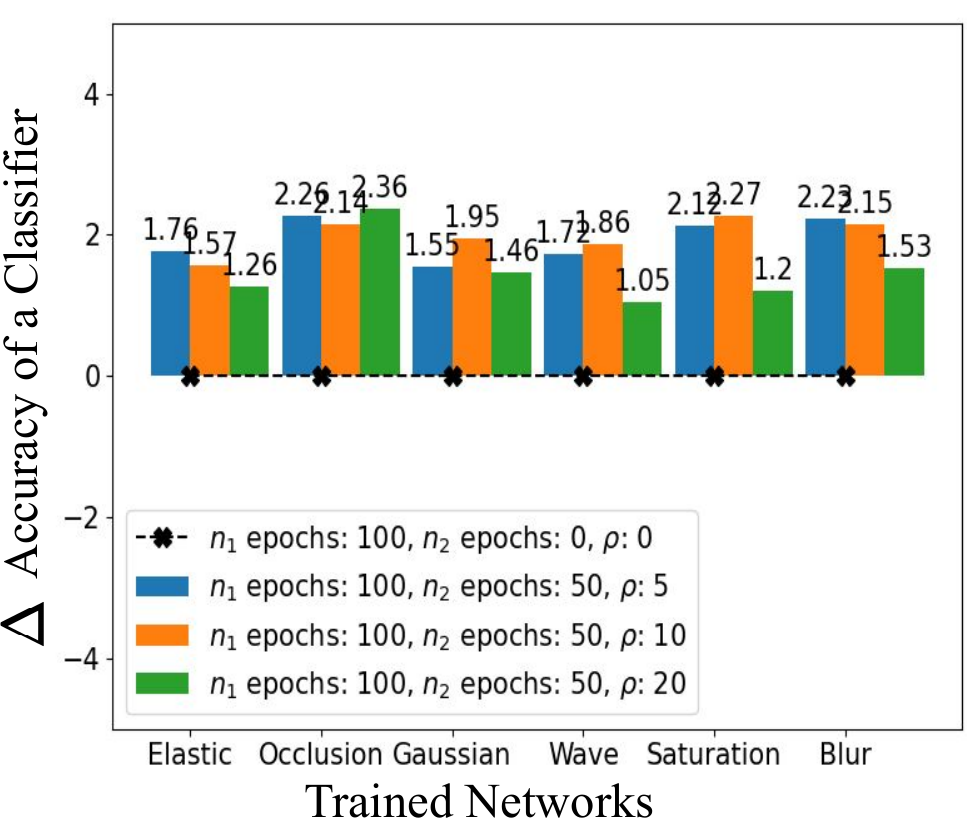}
         \caption{Evaluation by varying the drop $\rho$ to $5\%$, $10\%$ and $20 \%$ \\. }
         \label{fig:vary_drop}
     \end{subfigure}
        \caption{Ablation on Cifar-10 clean: Figure \ref{fig:vary_clean} shows training with perturbed images for a small number of epochs performs better than training with clean for a large number of epochs. Figure \ref{fig:vary_pert_epoch} shows perturbed training with an average number of epochs performs best. Figure \ref{fig:vary_drop} shows a moderate drop of $10\%$ leads to the best performance on clean images.  }
        \label{fig:ablation}
        \vspace{-0.3cm}
\end{figure*}

Figure \ref{fig:AT_adv} shows the results for adversarial images tested on adversarially robustified networks. We observe that adversarial training helps against adversarial perturbations. However, the recovery in the performance of natural perturbations with the natural perturbed training is higher for all datasets except Cifar-10. Hence, our natural perturbed training shows better generalization on perturbation in images seen during training as compared to adversarial training on seen adversarial perturbations.
\subsection{General Robustness: Evaluating Robustified Networks on Unseen Perturbations}
In Figure \ref{fig:unseen} we contrast the general robustness of natural perturbed training with adversarial training by testing them for unseen perturbations i.e. perturbations not seen during the training.

\myparagraph{Effectiveness of Natural Perturbed Training on Unseen Perturbations.} 
Figure \ref{fig:nat_elastic}, \ref{fig:nat_occ} shows the performance of elastic perturbed training and occlusion perturbed training tested on unseen adversarial and natural perturbations respectively. 
Results show that robustification with both elastic and occlusion perturbations recover the drop due to adversarial perturbations (plus symbol). We observe that natural perturbed training generalizes to other natural perturbations, except for elastic perturbed training on Gaussian noise for StanfordCars, AwA and LAD (red star symbol). Coarse grained Cifar-10 and STL-10 show the highest recovery on unseen natural perturbations. Hence, our natural perturbed training shows general robustness over adversarial as well as natural perturbations, while being even remarkable for coarse-grained datasets.  

\myparagraph{Ineffectiveness of Adversarial Training on Unseen Perturbations.} 
Figure \ref{fig:AT_nat} shows the results for an adversarially trained network (depicted by yellow symbols) and tested on unseen natural perturbations elastic (circle symbol) and occlusion (square symbol). Adversarial training does not generalize to unseen natural perturbations for fine grained datasets. It even leads to a further drop in the performance for them. For CUB and LAD, the drop almost doubles. For the coarse grained Cifar-10 dataset it helps against occlusion perturbation and for STL-10 it helps for all perturbations. However, the recovery is smaller than with the natural perturbed training. Hence, natural perturbed training shows better generalization than adversarial training for unseen perturbations.

\subsection{Ablation Studies} We perform ablation studies in Figure \ref{fig:ablation} on Cifar-10 by varying the parameters of the natural perturbed training. In Figure \ref{fig:vary_clean} we compare a standard network trained for 200 epochs on clean images with natural perturbed training by  100 clean and 50 perturbed epochs. Results show that natural perturbed training with a smaller number of epochs achieves better performance for clean images than a standard network with a larger number of epochs.

In Figure \ref{fig:vary_pert_epoch} we vary the number of perturbed epochs $n_2$ and test the performance of networks for clean images. Results depict that perturbed training with an average number of 50 epochs performs best. Figure \ref{fig:vary_drop} compares the performance of networks trained with different perturbation levels leading to drops of $5\%$, $10\%$ and $20 \%$. Results show that a moderate drop of $10 \%$ leads to the best performance on clean images. 

Finally, in Table \ref{table:multi} we evaluate the robustness of a network trained with multiple perturbations applied to the same image during subsequent $n_2$ epochs. The perturbations are elastic, occlusion, Gaussian noise and saturation. Compared to the results in Figure \ref{fig:nat_seen}, \ref{fig:nat_elastic} and \ref{fig:nat_occ} we observe an improvement in the recovery. Hence, training with multiple perturbations helps to enhance the overall robustness of the network.
\begin{table} 
\begin{center}
\resizebox{\linewidth}{!}{\begin{tabular}{|l | c | c | c |c| }
\hline 
Input:& Clean &Adversarial  &Elastic    &Occlusion    \\\hline
 $\bigtriangleup$:&
 1.63  &
 1.38  &
-3.14  &
 -0.62  \\\hline
 Input:& Gaussian Noise    & Wave   & Saturation   & Blur   \\\hline
  $\bigtriangleup$:&-1.67  & 
-5.12 &
 -0.17&
 -3.59 \\ \hline
\end{tabular}}
\end{center}
\vspace{-0.5cm}
\caption{Multiple perturbations training. $\bigtriangleup$ shows the change in the accuracy between a standard network and robustified one. Numbers in positive show an improvement in performance, negative show the drop not recovered from the initial $10\%$ drop. In contrast with the Cifar-10 results in Figure \ref{fig:nat_seen}, \ref{fig:nat_elastic} and \ref{fig:nat_occ} we observe a better generalization with multiple perturbations training.  }\label{table:multi}
\vspace{-0.5cm}
\end{table} 
\section{Conclusions}
A standardization procedure based on the effect of perturbations on the accuracy of the network is presented for fair quantitative evaluation of robustness. We introduced a new training procedure for enhancing the robustness of classifiers against perturbations. We demonstrated the effectiveness of our natural perturbed training for clean, adversarial and natural perturbations, both seen as well as unseen during the training. Our results showed that natural perturbed training, while being computationally fast, also shows better generalization on adversarial and natural perturbations than adversarial training. Moreover, it improves the classifier accuracy on clean images for the fine-grained CUB and StanfordCars, while for coarse-grained Cifar-10 and STL-10 improving the state of the art. \cite{sosnovik2019scale}.  
{\small
\bibliographystyle{ieee_fullname}
\bibliography{egbib}

\begin{thebibliography}{10}\itemsep=-1pt

\bibitem{azulay2018deep}
Aharon Azulay and Yair Weiss.
\newblock Why do deep convolutional networks generalize so poorly to small
  image transformations?
\newblock {\em arXiv preprint arXiv:1805.12177}, 2018.

\bibitem{carlini2017towards}
Nicholas Carlini and David Wagner.
\newblock Towards evaluating the robustness of neural networks.
\newblock In {\em 2017 ieee symposium on security and privacy (sp)}, pages
  39--57. IEEE, 2017.

\bibitem{coates2011analysis}
Adam Coates, Andrew Ng, and Honglak Lee.
\newblock An analysis of single-layer networks in unsupervised feature
  learning.
\newblock In {\em Proceedings of the fourteenth international conference on
  artificial intelligence and statistics}, pages 215--223. JMLR Workshop and
  Conference Proceedings, 2011.

\bibitem{5206848}
J. {Deng}, W. {Dong}, R. {Socher}, L. {Li}, {Kai Li}, and {Li Fei-Fei}.
\newblock Imagenet: A large-scale hierarchical image database.
\newblock In {\em 2009 IEEE Conference on Computer Vision and Pattern
  Recognition}, pages 248--255, 2009.

\bibitem{dodge2017study}
Samuel Dodge and Lina Karam.
\newblock A study and comparison of human and deep learning recognition
  performance under visual distortions.
\newblock In {\em 2017 26th international conference on computer communication
  and networks (ICCCN)}, pages 1--7. IEEE, 2017.

\bibitem{dong2020benchmarking}
Yinpeng Dong, Qi-An Fu, Xiao Yang, Tianyu Pang, Hang Su, Zihao Xiao, and Jun
  Zhu.
\newblock Benchmarking adversarial robustness on image classification.
\newblock In {\em Proceedings of the IEEE/CVF Conference on Computer Vision and
  Pattern Recognition}, pages 321--331, 2020.

\bibitem{engstrom2019exploring}
Logan Engstrom, Brandon Tran, Dimitris Tsipras, Ludwig Schmidt, and Aleksander
  Madry.
\newblock Exploring the landscape of spatial robustness.
\newblock In {\em International Conference on Machine Learning}, pages
  1802--1811, 2019.

\bibitem{fawzi2015manitest}
Alhussein Fawzi and Pascal Frossard.
\newblock Manitest: Are classifiers really invariant?
\newblock {\em arXiv preprint arXiv:1507.06535}, 2015.

\bibitem{ford2019adversarial}
Nic Ford, Justin Gilmer, Nicolas Carlini, and Dogus Cubuk.
\newblock Adversarial examples are a natural consequence of test error in
  noise.
\newblock {\em arXiv preprint arXiv:1901.10513}, 2019.

\bibitem{geirhos2017comparing}
Robert Geirhos, David~HJ Janssen, Heiko~H Sch{\"u}tt, Jonas Rauber, Matthias
  Bethge, and Felix~A Wichmann.
\newblock Comparing deep neural networks against humans: object recognition
  when the signal gets weaker.
\newblock {\em arXiv preprint arXiv:1706.06969}, 2017.

\bibitem{goodfellow2014explaining}
Ian~J Goodfellow, Jonathon Shlens, and Christian Szegedy.
\newblock Explaining and harnessing adversarial examples.
\newblock {\em arXiv preprint arXiv:1412.6572}, 2014.

\bibitem{hendrycks2020many}
Dan Hendrycks, Steven Basart, Norman Mu, Saurav Kadavath, Frank Wang, Evan
  Dorundo, Rahul Desai, Tyler Zhu, Samyak Parajuli, Mike Guo, et~al.
\newblock The many faces of robustness: A critical analysis of
  out-of-distribution generalization.
\newblock {\em arXiv preprint arXiv:2006.16241}, 2020.

\bibitem{hendrycks2019benchmarking}
Dan Hendrycks and Thomas Dietterich.
\newblock Benchmarking neural network robustness to common corruptions and
  perturbations.
\newblock {\em arXiv preprint arXiv:1903.12261}, 2019.

\bibitem{hendrycks2019using}
Dan Hendrycks, Kimin Lee, and Mantas Mazeika.
\newblock Using pre-training can improve model robustness and uncertainty.
\newblock In {\em International Conference on Machine Learning}, pages
  2712--2721. PMLR, 2019.

\bibitem{kanbak2018geometric}
Can Kanbak, Seyed-Mohsen Moosavi-Dezfooli, and Pascal Frossard.
\newblock Geometric robustness of deep networks: analysis and improvement.
\newblock In {\em Proceedings of the IEEE Conference on Computer Vision and
  Pattern Recognition}, pages 4441--4449, 2018.

\bibitem{KrauseStarkDengFei-Fei_3DRR2013}
Jonathan Krause, Michael Stark, Jia Deng, and Li Fei-Fei.
\newblock 3d object representations for fine-grained categorization.
\newblock In {\em 4th International IEEE Workshop on 3D Representation and
  Recognition (3dRR-13)}, Sydney, Australia, 2013.

\bibitem{krizhevsky2009learning}
Alex Krizhevsky et~al.
\newblock Learning multiple layers of features from tiny images.
\newblock 2009.

\bibitem{kurakin2016adversarial}
Alexey Kurakin, Ian Goodfellow, and Samy Bengio.
\newblock Adversarial examples in the physical world.
\newblock {\em arXiv preprint arXiv:1607.02533}, 2016.

\bibitem{laugros2019adversarial}
Alfred Laugros, Alice Caplier, and Matthieu Ospici.
\newblock Are adversarial robustness and common perturbation robustness
  independant attributes?
\newblock In {\em Proceedings of the IEEE International Conference on Computer
  Vision Workshops}, pages 0--0, 2019.

\bibitem{laugros2021increasing}
Alfred LAUGROS, Alice Caplier, and Matthieu Ospici.
\newblock Increasing the coverage and balance of robustness benchmarks by using
  non-overlapping corruptions, 2021.

\bibitem{madry2017towards}
Aleksander Madry, Aleksandar Makelov, Ludwig Schmidt, Dimitris Tsipras, and
  Adrian Vladu.
\newblock Towards deep learning models resistant to adversarial attacks.
\newblock {\em arXiv preprint arXiv:1706.06083}, 2017.

\bibitem{miyato2018virtual}
Takeru Miyato, Shin-ichi Maeda, Masanori Koyama, and Shin Ishii.
\newblock Virtual adversarial training: a regularization method for supervised
  and semi-supervised learning.
\newblock {\em IEEE transactions on pattern analysis and machine intelligence},
  41(8):1979--1993, 2018.

\bibitem{moosavi2016deepfool}
Seyed-Mohsen Moosavi-Dezfooli, Alhussein Fawzi, and Pascal Frossard.
\newblock Deepfool: a simple and accurate method to fool deep neural networks.
\newblock In {\em CVPR}, 2016.

\bibitem{papernot2016limitations}
Nicolas Papernot, Patrick McDaniel, Somesh Jha, Matt Fredrikson, Z~Berkay
  Celik, and Ananthram Swami.
\newblock The limitations of deep learning in adversarial settings.
\newblock In {\em EuroS\&P}. IEEE, 2016.

\bibitem{recht2018cifar}
Benjamin Recht, Rebecca Roelofs, Ludwig Schmidt, and Vaishaal Shankar.
\newblock Do cifar-10 classifiers generalize to cifar-10?
\newblock {\em arXiv preprint arXiv:1806.00451}, 2018.

\bibitem{rusak2020increasing}
Evgenia Rusak, Lukas Schott, Roland Zimmermann, Julian Bitterwolf, Oliver
  Bringmann, Matthias Bethge, and Wieland Brendel.
\newblock Increasing the robustness of dnns against image corruptions by
  playing the game of noise.
\newblock {\em arXiv preprint arXiv:2001.06057}, 2020.

\bibitem{schneider2020improving}
Steffen Schneider, Evgenia Rusak, Luisa Eck, Oliver Bringmann, Wieland Brendel,
  and Matthias Bethge.
\newblock Improving robustness against common corruptions by covariate shift
  adaptation.
\newblock {\em arXiv preprint arXiv:2006.16971}, 2020.

\bibitem{simard2003best}
Patrice~Y Simard, David Steinkraus, John~C Platt, et~al.
\newblock Best practices for convolutional neural networks applied to visual
  document analysis.
\newblock Citeseer.

\bibitem{song2019robust}
Chubiao Song, Kun He, Jiadong Lin, Liwei Wang, and John~E Hopcroft.
\newblock Robust local features for improving the generalization of adversarial
  training.
\newblock {\em arXiv preprint arXiv:1909.10147}, 2019.

\bibitem{sosnovik2019scale}
Ivan Sosnovik, Micha{\l} Szmaja, and Arnold Smeulders.
\newblock Scale-equivariant steerable networks.
\newblock {\em arXiv preprint arXiv:1910.11093}, 2019.

\bibitem{su2018robustness}
Dong Su, Huan Zhang, Hongge Chen, Jinfeng Yi, Pin-Yu Chen, and Yupeng Gao.
\newblock Is robustness the cost of accuracy?--a comprehensive study on the
  robustness of 18 deep image classification models.
\newblock In {\em Proceedings of the European Conference on Computer Vision
  (ECCV)}, pages 631--648, 2018.

\bibitem{su2019one}
Jiawei Su, Danilo~Vasconcellos Vargas, and Kouichi Sakurai.
\newblock One pixel attack for fooling deep neural networks.
\newblock {\em TEVC}, 2019.

\bibitem{szegedy2013intriguing}
Christian Szegedy, Wojciech Zaremba, Ilya Sutskever, Joan Bruna, Dumitru Erhan,
  Ian Goodfellow, and Rob Fergus.
\newblock Intriguing properties of neural networks.
\newblock {\em ICLR}, 2013.

\bibitem{tang2021selfnorm}
Zhiqiang Tang, Yunhe Gao, Yi Zhu, Zhi Zhang, Mu Li, and Dimitris~N. Metaxas.
\newblock Selfnorm and crossnorm for out-of-distribution robustness, 2021.

\bibitem{tsipras2018robustness}
Dimitris Tsipras, Shibani Santurkar, Logan Engstrom, Alexander Turner, and
  Aleksander Madry.
\newblock Robustness may be at odds with accuracy.
\newblock {\em arXiv preprint arXiv:1805.12152}, 2018.

\bibitem{WelinderEtal2010}
P. Welinder, S. Branson, T. Mita, C. Wah, F. Schroff, S. Belongie, and P.
  Perona.
\newblock {Caltech-UCSD Birds 200}.
\newblock Technical Report CNS-TR-2010-001, California Institute of Technology,
  2010.

\bibitem{wong2020fast}
Eric Wong, Leslie Rice, and J~Zico Kolter.
\newblock Fast is better than free: Revisiting adversarial training.
\newblock {\em arXiv preprint arXiv:2001.03994}, 2020.

\bibitem{8413121}
Y. {Xian}, C.~H. {Lampert}, B. {Schiele}, and Z. {Akata}.
\newblock Zero-shot learning—a comprehensive evaluation of the good, the bad
  and the ugly.
\newblock {\em IEEE Transactions on Pattern Analysis and Machine Intelligence},
  41(9):2251--2265, 2019.

\bibitem{yin2019fourier}
Dong Yin, Raphael~Gontijo Lopes, Jonathon Shlens, Ekin~D Cubuk, and Justin
  Gilmer.
\newblock A fourier perspective on model robustness in computer vision.
\newblock {\em arXiv preprint arXiv:1906.08988}, 2019.

\bibitem{zhang2019theoretically}
Hongyang Zhang, Yaodong Yu, Jiantao Jiao, Eric~P Xing, Laurent~El Ghaoui, and
  Michael~I Jordan.
\newblock Theoretically principled trade-off between robustness and accuracy.
\newblock {\em arXiv preprint arXiv:1901.08573}, 2019.

\bibitem{zhang2019interpreting}
Tianyuan Zhang and Zhanxing Zhu.
\newblock Interpreting adversarially trained convolutional neural networks.
\newblock {\em arXiv preprint arXiv:1905.09797}, 2019.

\bibitem{zhao2018large}
Bo Zhao, Yanwei Fu, Rui Liang, Jiahong Wu, Yonggang Wang, and Yizhou Wang.
\newblock A large-scale attribute dataset for zero-shot learning.
\newblock {\em arXiv preprint arXiv:1804.04314}, 2018.

\end{thebibliography}
}
\clearpage
\begin{figure*}[t]
     \centering
     \begin{subfigure}[b]{0.31\textwidth}
         \centering
         \includegraphics[width=\textwidth]{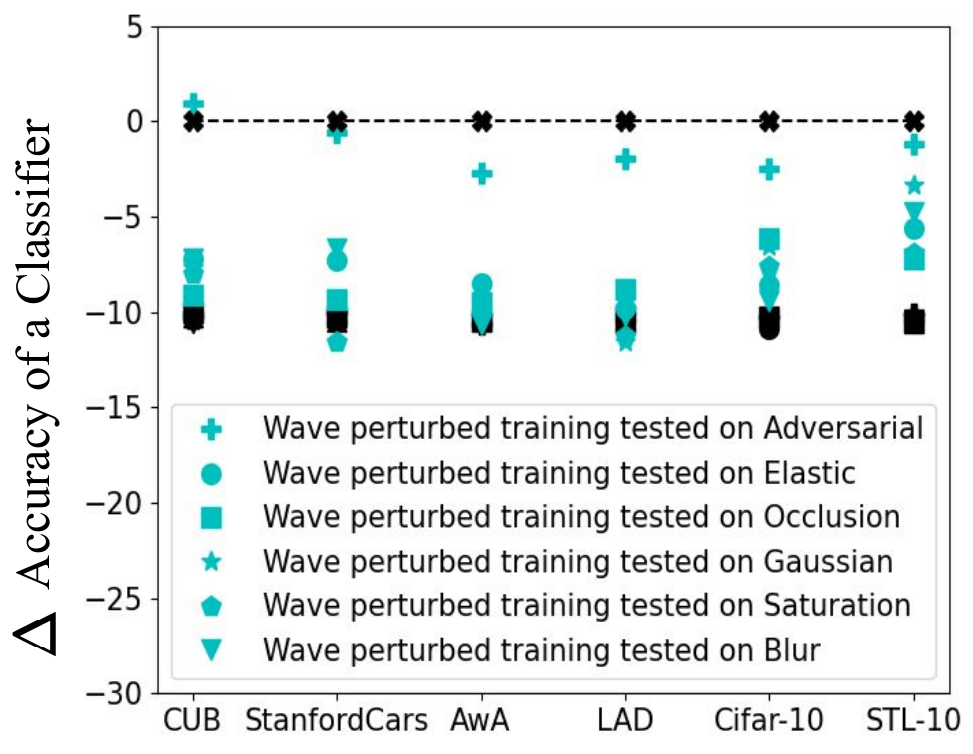}
         \caption{Evaluating wave perturbed training for unseen natural perturbations.  }
         \label{fig:nat_wave}
     \end{subfigure}
     \hspace{1em}%
     \begin{subfigure}[b]{0.31\textwidth}
         \centering
         \includegraphics[width=\textwidth]{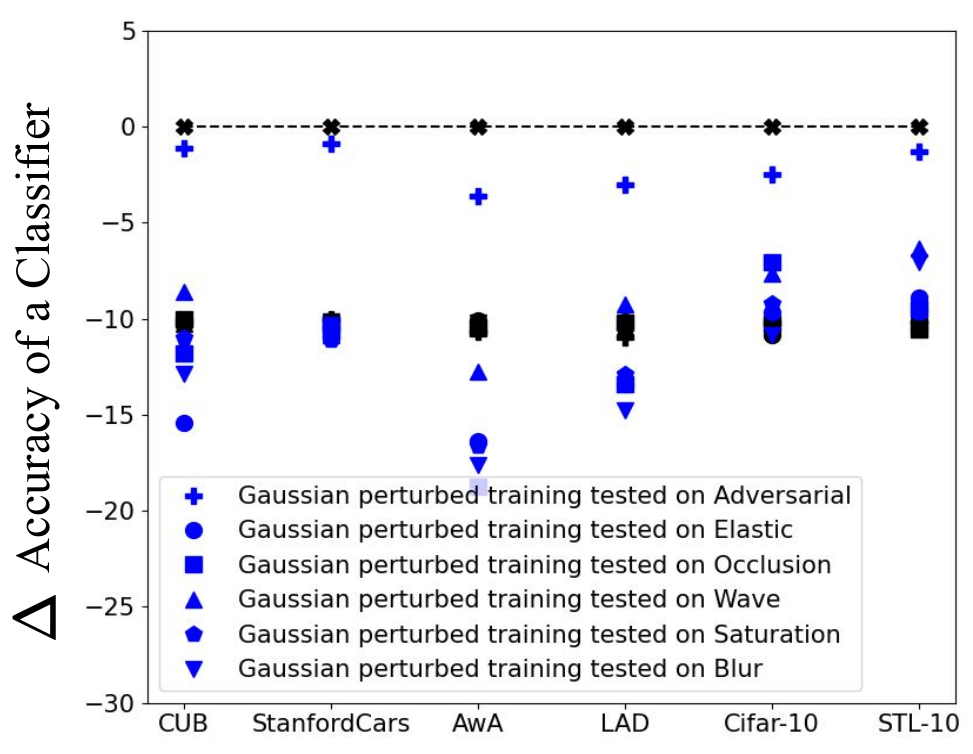}
         \caption{Evaluating Gaussian perturbed training for unseen natural perturbations.  }
         \label{fig:nat_gaus}
     \end{subfigure}
     \hspace{1em}%
     \begin{subfigure}[b]{0.31\textwidth}
         \centering
         \includegraphics[width=\textwidth]{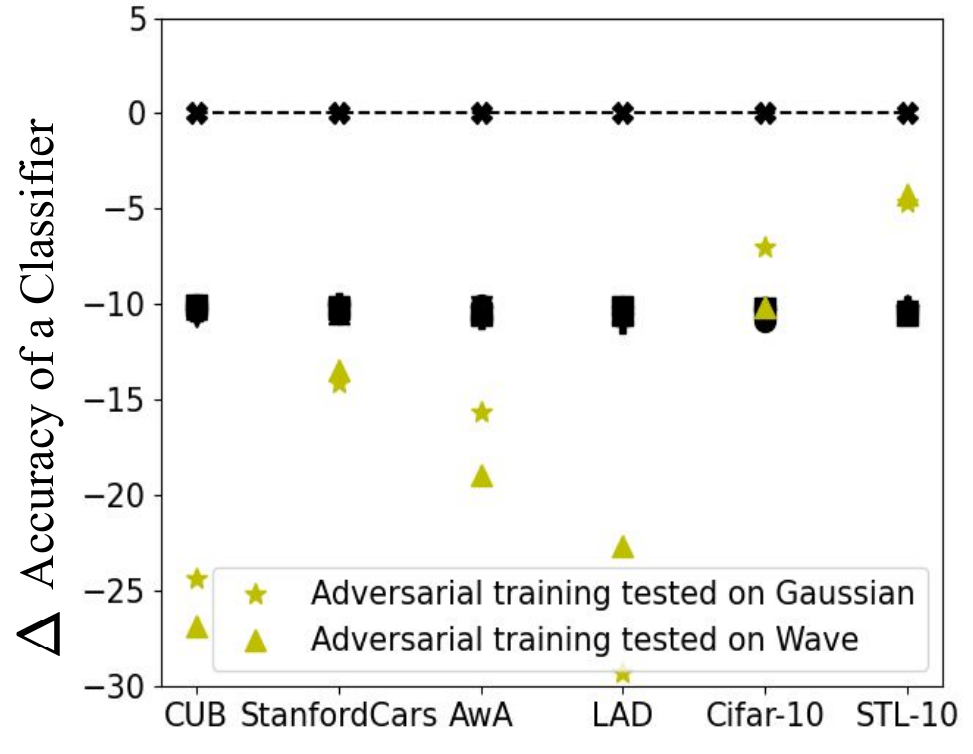}
         \caption{{Evaluating Adversarial training for unseen natural perturbations.  } }
         \label{fig:AT_nat_1}
     \end{subfigure}
        \caption{Comparing the performance of Natural perturbed (wave, Gaussian) training with Adversarial training on \textit{unseen perturbations}. The type of symbol represents test perturbation and color of the symbol represents the type of training perturbation. Adversarial training shows some general robustness on coarse-grained datasets but for fine-grained datasets it fails to generalize. Natural perturbed training generalizes to adversarial perturbations (plus symbol) and other natural perturbations.}
        \label{fig:unseen-1}
        \vspace{-0.2cm}
\end{figure*}
\begin{figure*}[t]
     \centering
     \begin{subfigure}[b]{0.31\textwidth}
         \centering
         \includegraphics[width=\textwidth]{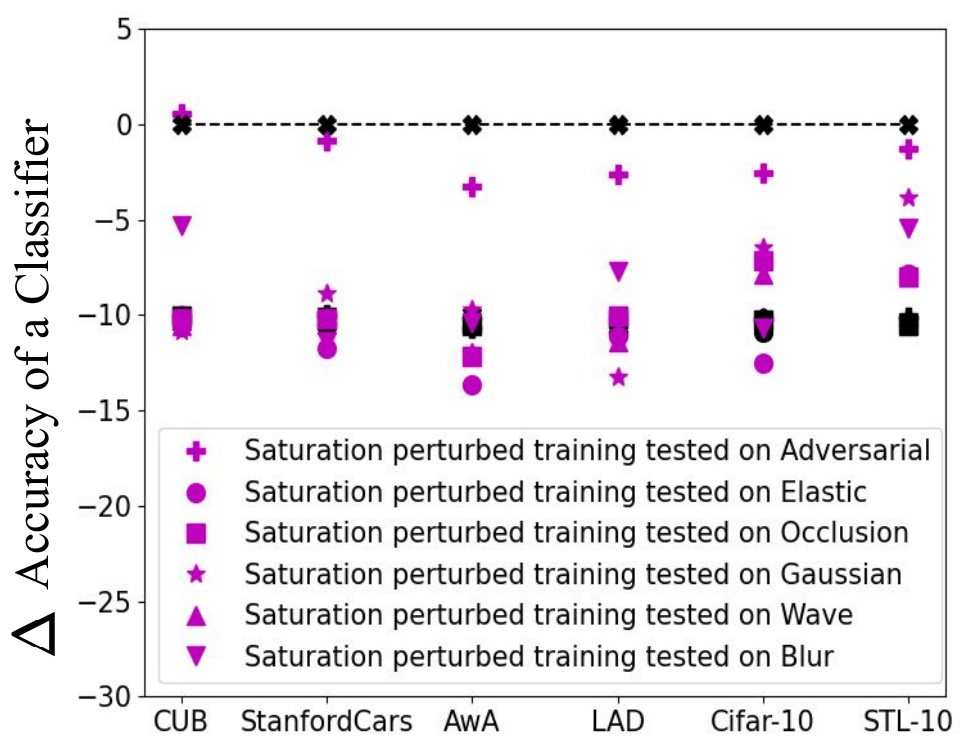}
         \caption{Evaluating saturation perturbed training for unseen natural perturbations.  }
         \label{fig:nat_sat}
     \end{subfigure}
     \hspace{1em}%
     \begin{subfigure}[b]{0.31\textwidth}
         \centering
         \includegraphics[width=\textwidth]{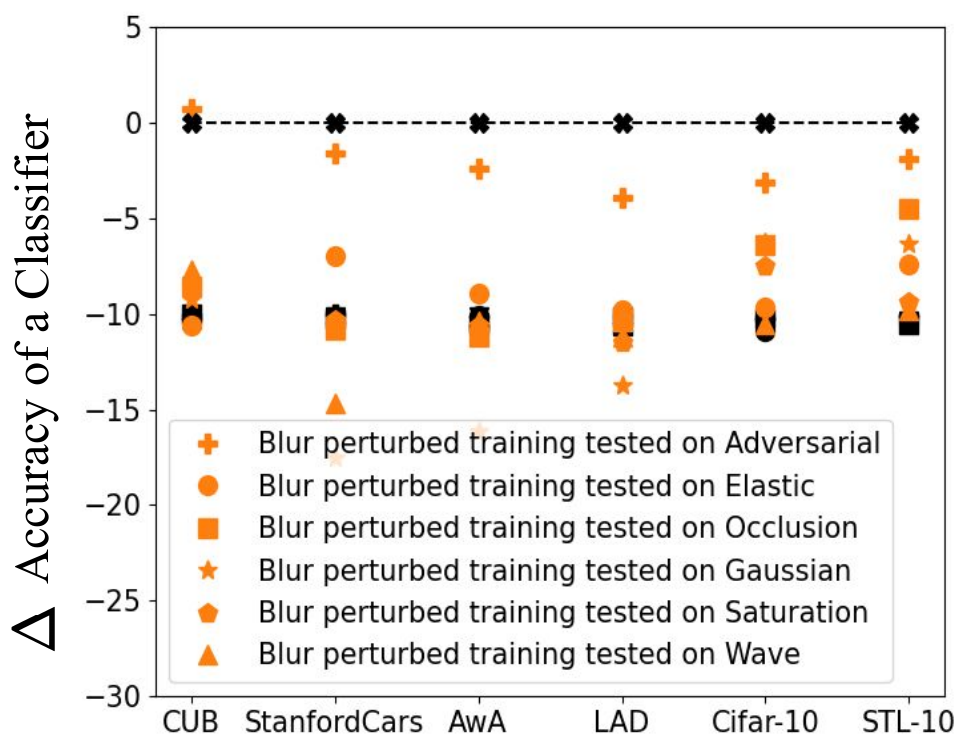}
         \caption{Evaluating Gaussian blur perturbed training for unseen natural perturbations.  }
         \label{fig:nat_gaus_blur}
     \end{subfigure}
     \hspace{1em}%
     \begin{subfigure}[b]{0.31\textwidth}
         \centering
         \includegraphics[width=\textwidth]{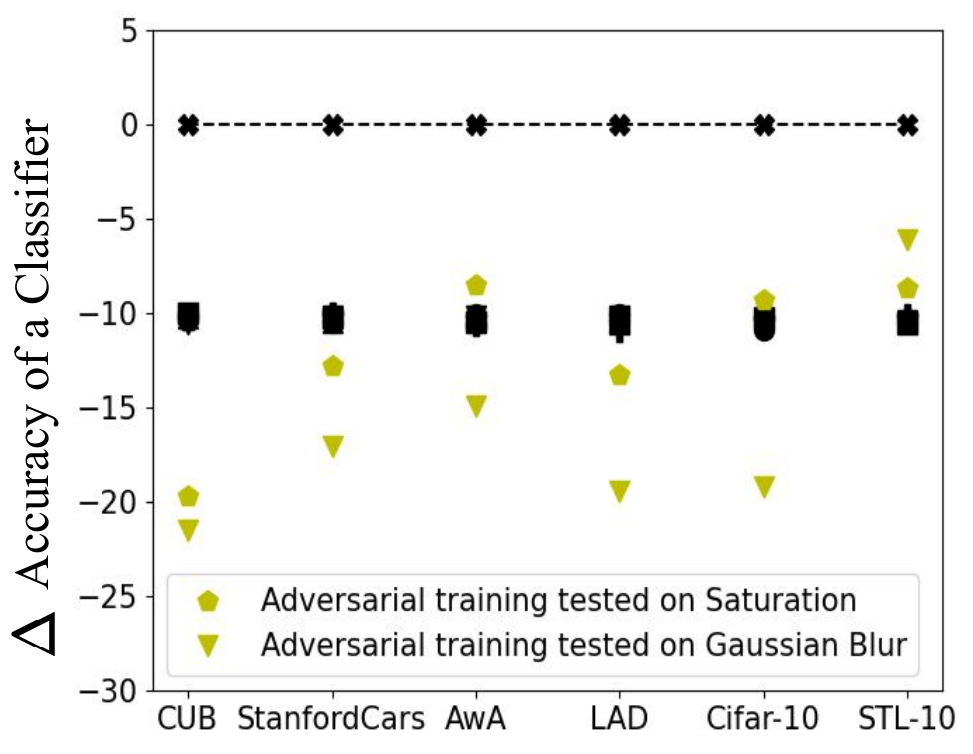}
         \caption{{Evaluating Adversarial training for unseen natural perturbations.  } }
         \label{fig:AT_nat_2}
     \end{subfigure}
        \caption{Comparing the performance of Natural perturbed (saturation, Gaussian blur) training with Adversarial training on \textit{unseen perturbations}. The type of symbol represents test perturbation and color of the symbol represents the type of training perturbation. Adversarial training shows some general robustness on coarse-grained datasets but for fine-grained datasets it fails to generalize. Natural perturbed training generalizes to adversarial perturbations (plus symbol) and other natural perturbations.}
        \label{fig:unseen-2}
        \vspace{-0.2cm}
\end{figure*}
\section{Supplementary Material}
\subsection{General Robustness of Wave and Gaussian Perturbations: Evaluating Robustified Networks on Unseen Perturbations}
In Figure \ref{fig:unseen-1} we contrast the general robustness of natural perturbed training with adversarial training by testing them for unseen perturbations i.e. perturbations not seen during the training. Here, we present the results for wave and Gaussian noise.

\myparagraph{Effectiveness of Natural Perturbed Training on Unseen Perturbations.} 
Figure \ref{fig:nat_wave}, \ref{fig:nat_gaus} shows the performance of wave perturbed training and Gaussian perturbed training tested on unseen adversarial and natural perturbations respectively. 
Results show that robustification with both wave and Gaussian perturbations recover the drop due to adversarial perturbations (plus symbol). We observe that natural perturbed training especially wave perturbed training generalizes to other natural perturbations too. Gaussian perturbed training also generalizes to other natural perturbations except for CUB, AwA and LAD datasets. Coarse grained Cifar-10 and STL-10 show the highest recovery on unseen natural perturbations. The recovery for coarse grained datasets with wave perturbations is better than the Gaussian noise. Hence, our natural perturbed training shows general robustness over adversarial as well as natural perturbations, while being even remarkable for coarse-grained datasets.  

\myparagraph{Ineffectiveness of Adversarial Training on Unseen Wave and Gaussian Perturbations.} 
Figure \ref{fig:AT_nat_1} shows the results for an adversarially trained network (depicted by yellow symbols) and tested on unseen natural perturbations wave (triangle symbol) and Gaussian (star symbol). Adversarial training does not generalize to unseen natural perturbations for fine grained datasets. It even leads to a further drop in the performance for them. For CUB and LAD, the drop almost triples. For the coarse grained Cifar-10 dataset it helps against Gaussian perturbation and for STL-10 it helps for both perturbations. However, the recovery is smaller than with the natural perturbed training. Hence, natural perturbed training shows better generalization than adversarial training for unseen perturbations.

\subsection{General Robustness of Saturation and Gaussian Blur Perturbations: Evaluating Robustified Networks on Unseen Perturbations}
In Figure \ref{fig:unseen-2} we compare the general robustness of natural perturbed training with adversarial training by testing them for unseen perturbations i.e. perturbations not seen during the training. Here, we present results for saturation and Gaussian blur.

\myparagraph{Effectiveness of Natural Perturbed Training on Unseen Perturbations.} 
Figure \ref{fig:nat_sat}, \ref{fig:nat_gaus_blur} shows the performance of saturation perturbed training and Gaussian blur perturbed training tested on unseen adversarial and natural perturbations respectively. 
Results show that robustification with both saturation and Gaussian blur perturbations recover the drop due to adversarial perturbations (plus symbol). We observe that natural perturbed training generalizes to other natural perturbations too. Except for saturation perturbed on AwA and LAD datasets. Coarse grained Cifar-10 and STL-10 show the highest recovery on unseen natural perturbations. Hence, our natural perturbed training shows general robustness over adversarial as well as natural perturbations, while being even noteworthy for coarse-grained datasets.  

\myparagraph{Ineffectiveness of Adversarial Training on Unseen wave and Gaussian Perturbations.} 
Figure \ref{fig:AT_nat_2} shows the results for an adversarially trained network (depicted by yellow symbols) and tested on unseen natural perturbations saturation (five pointed star symbol) and Gaussian blur (triangle down symbol). Adversarial training does not generalize to unseen natural perturbations for fine grained datasets. It even leads to a further drop in the performance for them. For CUB, LAD and Cifar-10 Gaussian blur test set the drop almost doubles. For coarse grained Cifar-10 saturation test it neither helps nor degrades the performance. For STL-10 it helps for both perturbations. However, the recovery is smaller than with the natural perturbed training. Hence, natural perturbed training shows better generalization than adversarial training for unseen perturbations.
\end{document}